\documentclass[10pt,twocolumn,letterpaper]{article}
\usepackage[accsupp]{axessibility} 

\usepackage[applications]{wacv}      

\usepackage{graphicx}
\usepackage{amsmath}
\usepackage{amssymb}
\usepackage{booktabs}

\usepackage[table]{xcolor}
\usepackage{colortbl}
\definecolor{acceptgreen}{rgb}{0, 0.59, 0.24}
\definecolor{rejectred}{rgb}{0.94, 0.11, 0.20}

\colorlet{algoblue}{cyan!60}
\colorlet{algored}{red!40}
\colorlet{algogreen}{green!60}

\usepackage{bbding}

\usepackage{algorithm}
\usepackage{algorithmic}  

\usepackage{tikz}
\usetikzlibrary{fit,calc}
\newcommand*{\tikzmk}[1]{\tikz[remember picture,overlay,] \node (#1) {};\ignorespaces}
\newcommand{\boxit}[1]{\tikz[remember picture,overlay]{\node[xshift=5pt, yshift=3pt,fill=#1,opacity=.10,fit={(begin)($(end)+(0, -0.3\baselineskip)$)}] {};}}

\newcommand {\first}[1]{\textcolor{red}{\textbf{#1}}}
\newcommand {\second}[1]{\textcolor{blue}{\underline{#1}}}

\usepackage[pagebackref,breaklinks,colorlinks]{hyperref}

\usepackage[capitalize]{cleveref}
\crefname{section}{Sec.}{Secs.}
\Crefname{section}{Section}{Sections}
\Crefname{table}{Table}{Tables}
\crefname{table}{Tab.}{Tabs.}

\def\wacvPaperID{1929} 
\def\confName{WACV}
\def\confYear{2025}

\begin{document}

\title{DiffQRCoder: Diffusion-based Aesthetic QR Code Generation with Scanning Robustness Guided Iterative Refinement}

\author{
Jia-Wei Liao$^{1, 2}$
\quad
Winston Wang$^{1}$\thanks{Equal contributions.}
\quad
Tzu-Sian Wang$^{1*}$
\quad
Li-Xuan Peng$^{1*}$
\quad
Ju-Hsuan Weng$^{1, 2}$ \\
\quad
Cheng-Fu Chou$^{2}$
\qquad
Jun-Cheng Chen$^{1}$
\vspace{0.5em} \\
$^{1}$ Research Center for Information Technology Innovation, Academia Sinica, \\
$^{2}$ National Taiwan University
}

\twocolumn[{
    \renewcommand\twocolumn[1][]{#1}%
    \maketitle   
    \begin{center}
        \newcommand{\teaserwidth}{\textwidth}
        \vspace{-0.15in}
        \centerline{
        \includegraphics[width=\teaserwidth, clip]{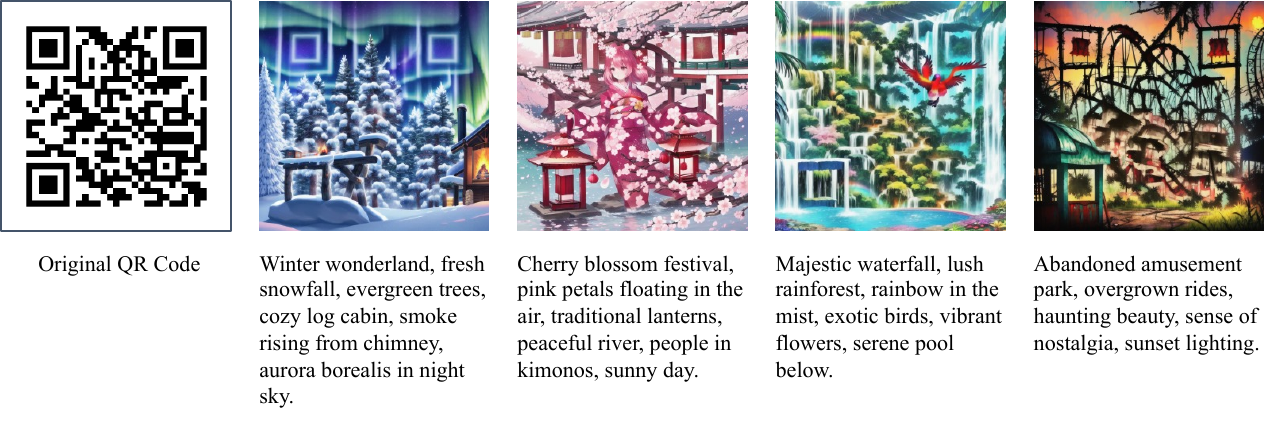}
        }
        \vspace{-2ex}
        \captionof{figure}{\textbf{Aesthetic QR codes generated from DiffQRCoder}. Our method takes a QR code and a text prompt as input to generate an aesthetic QR code. We leverage the pre-trained ControlNet and guide the generation process using our proposed Scanning Robust Perceptual Guidance (SRPG) to ensure the generated code is both scannable and attractive.}
    \label{fig:teaser}
    \vspace{-12pt}
    \end{center}
}]

\maketitle
\footnotetext{* denotes equal contribution.}

\begin{abstract}
With the success of Diffusion Models for image generation, the technologies also have revolutionized the aesthetic Quick Response (QR) code generation. Despite significant improvements in visual attractiveness for the beautified codes, their scannabilities are usually sacrificed and thus hinder their practical uses in real-world scenarios. To address this issue, we propose a novel training-free \textbf{Diff}usion-based \textbf{QR} \textbf{Code} generato\textbf{r} (DiffQRCoder) to effectively craft both scannable and visually pleasing QR codes. The proposed approach introduces Scanning-Robust Perceptual Guidance (SRPG), a new diffusion guidance for Diffusion Models to guarantee the generated aesthetic codes to obey the ground-truth QR codes while maintaining their attractiveness during the denoising process. Additionally, we present another post-processing technique, Scanning Robust Manifold Projected Gradient Descent (SR-MPGD), to further enhance their scanning robustness through iterative latent space optimization. With extensive experiments, the results demonstrate that our approach not only outperforms other compared methods in Scanning Success Rate (SSR) with better or comparable CLIP aesthetic score (CLIP-aes.) but also significantly improves the SSR of the ControlNet-only approach from 60\% to 99\%. The subjective evaluation indicates that our approach achieves promising visual attractiveness to users as well. Finally, even with different scanning angles and the most rigorous error tolerance settings, our approach robustly achieves over 95\% SSR, demonstrating its capability for real-world applications. Our project page is available at \url{https://jwliao1209.github.io/DiffQRCoder}.
\end{abstract}

\section{Introduction}
\label{sec:introduction}

\begin{figure}[t]
    \centering
    \includegraphics[width=0.75\linewidth]{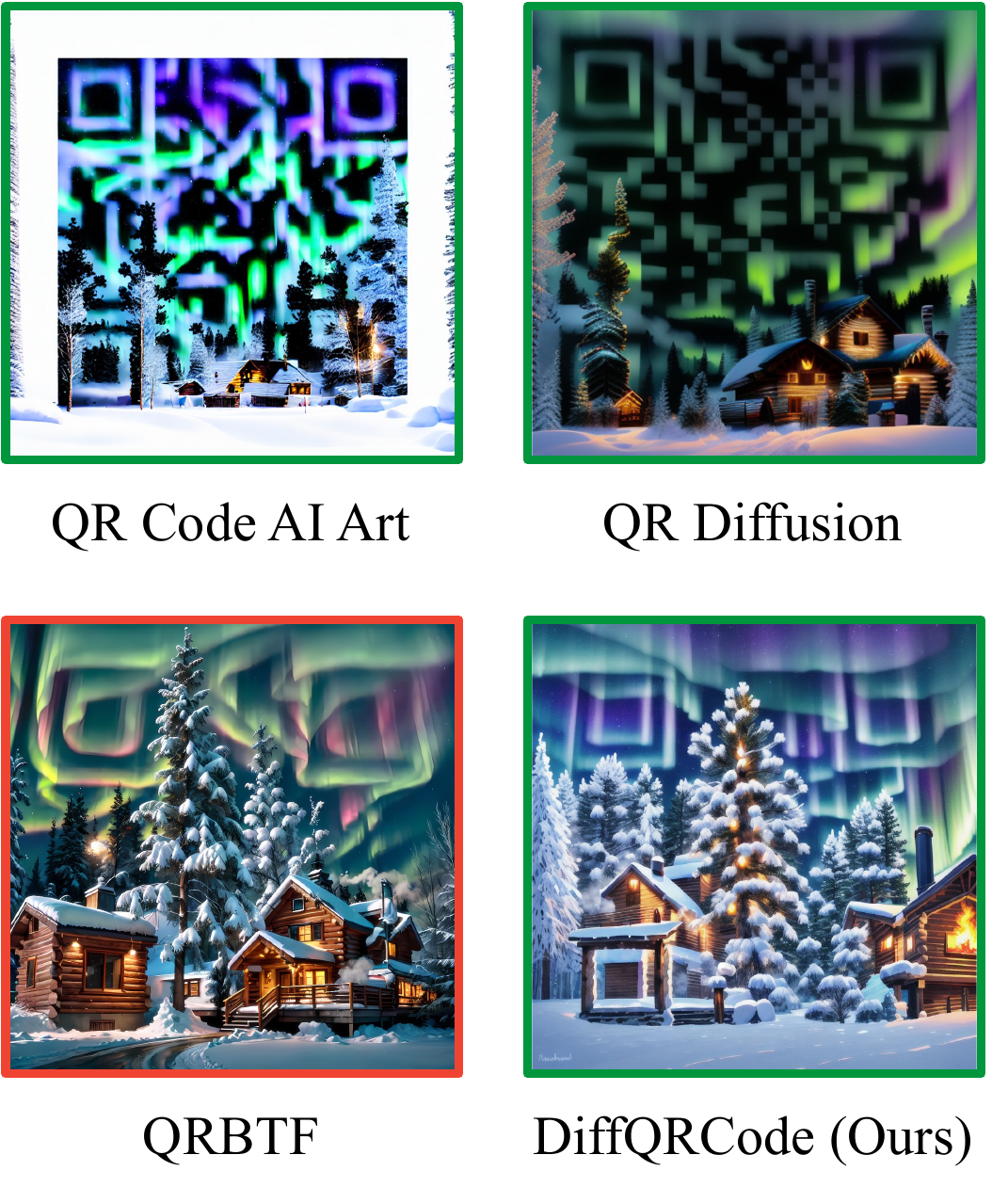}
    \vspace{-5pt}
    \caption{\textbf{Existing methods struggle to balance scannability and aesthetics.} Although QRBTF~\cite{qrbtf2023} generate visually appealing QR codes, they lack scanning robustness. Conversely, QR Code AI Art~\cite{qrcodeai} and QR Diffusion~\cite{qrdiffusion} produce better scanning robust QR codes but are visually less appealing. Our approach can generate both attractive and scannable QR codes. \textcolor{rejectred}{Red} frames indicate unscannable codes, while \textcolor{acceptgreen}{green} frames denote scannable codes. Zoom in for better viewing details.}
    \label{fig:dilemma}
    \vspace{-10pt}
\end{figure}

Quick Response (QR)~\cite{iso2000qr} codes are ubiquitous in daily transactions, information sharing, and marketing, driven by their quick readability and the widespread use of smartphones. However, the standard black-and-white QR codes lack visual appeal. Aesthetic QR codes offer a solution by not only capturing user attention but also seamlessly integrating with product designs, enhancing user experiences, and amplifying marketing effectiveness. By creating visually appealing QR codes, businesses can elevate brand engagement and improve advertising impact, making them a valuable tool for both functionality and design. Recognizing the commercial value of aesthetic QR codes, numerous beautification techniques have been thus developed.

For this purpose, some previous works have attempted to generate aesthetic QR codes via style-transfer-based techniques~\cite{su2021artcoder} to blend style textures with the QR code patterns. However, these methods often lack flexibility and can reduce scanning robustness.

Instead, current prevailing commercial products~\cite{qrcodemonster2023} have adopted generative models to create stylized QR codes, primarily employing Diffusion Models (e.g., ControlNet~\cite{zavadski2023controlnet}). The mainstream methodology is to adjust the Classifier-Free Guidance (CFG) weights~\cite{dhariwal2021diffusion} in ControlNet to create visually pleasing QR codes. However, selecting CFG weights presents a trade-off between scannability and visual quality (Fig.~\ref{fig:dilemma}). In practical applications, manual post-processing is often used to fix unscannable codes, but this process is time-consuming and labor-intensive. Therefore, it is still an open challenge to generate aesthetic QR codes with a good balance between visual attractiveness and scanning robustness.

To address the instability of scanability in previous generative-based methods, we propose \textbf{Diff}usion-based \textbf{QR} \textbf{Code} generato\textbf{r} (DiffQRCoder), a training-free approach to balance the scannability and aesthetics of QR codes. We introduce Scanning Robust Loss (SRL), specifically designed for evaluating the scannability of a beautified QR code with respect to its reference code.
Building on SRL, we develop Scanning Robust Perceptual Guidance (SRPG), an extension of the Classifier Guidance concept~\cite{dhariwal2021diffusion, yu2023freedom, bansal2024universal}, which ensures generation fidelity to ground-truth QR codes while preserving aesthetics during the denoising process.

Besides, we develop a post-processing technique called Scanning Robust Manifold Projected Gradient Descent (SR-MPGD) further to enhance the scanning robustness through iterative latent space optimization utilizing a pre-trained Variational Autoencoder (VAE)~\cite{kingma2014autoencoding}. Specifically, our framework features the following key designs: 1) The proposed framework is training-free and compatible with existing diffusion models. 2) Our approach exploits the error tolerance capability inherent in standard QR codes for more flexible and precise manipulation over QR code image pixels.

Finally, extensive experiments demonstrate that our approach outperforms other compared models in Scanning Success Rate (SSR) with better or comparable CLIP aesthetic scores (CLIP-aes.)~\cite{schuhmann2022laion}. Specifically, it achieves 99\% SSR while better preserving CLIP aesthetic scores.

Our main contributions are summarized as follows:
\begin{enumerate}
\item We propose a two-stage iterative refinement framework with a novel Scanning Robust Perceptual Guidance (SRPG) tailored for QR code mechanisms to generate scanning-robust and visually appealing aesthetic QR codes without training.

\item We propose Scanning Robust Manifold Projected Gradient Descent (SR-MPGD) for post-processing, enabling Scanning Success Rate (SSR) of aesthetic QR code up to 100\% through latent space optimization.

\item Extensive quantitative and qualitative experiments demonstrate that our proposed framework significantly enhances the Scanning Success Rate (SSR) of the ControlNet-only approach from 60\% to nearly 100\%, without compromising aesthetics. User subjective evaluations further confirm the visual appeal of our QR codes.
\end{enumerate}

\section{Related Works}

\begin{figure*}[t]
    \centering
    \includegraphics[width=\textwidth]{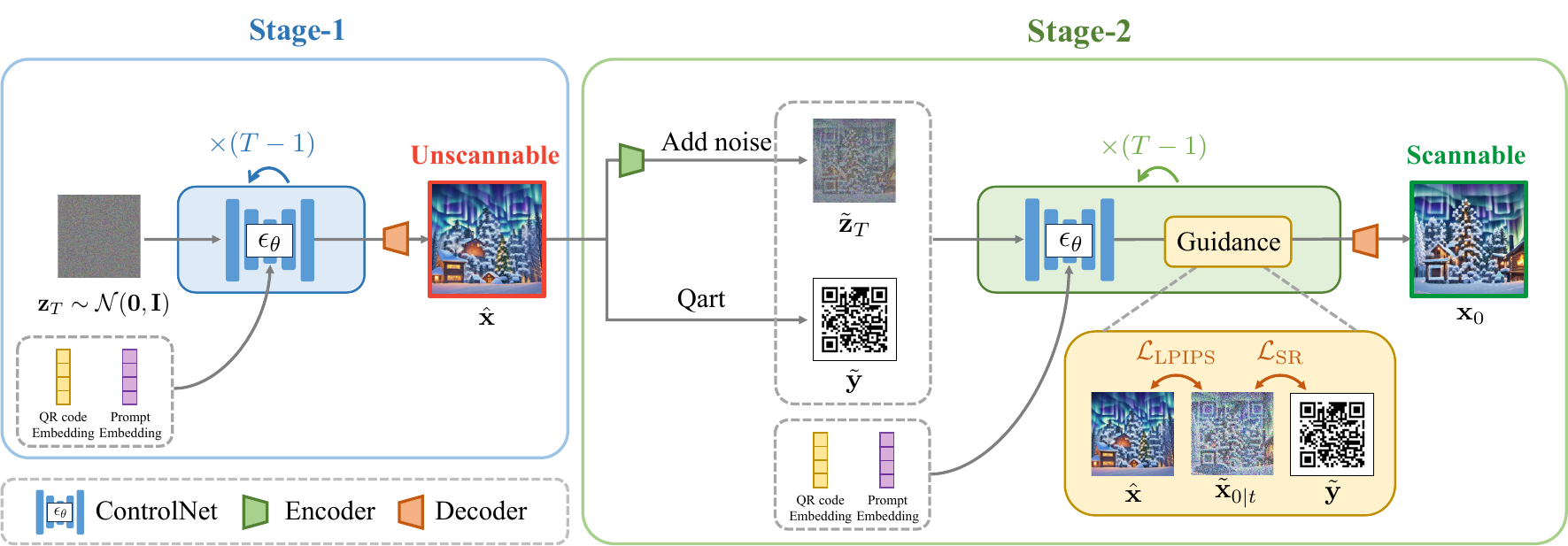}
    \vspace{-15pt}
    \caption{\textbf{An overview of our two-stage pipeline with Scanning Robust Perceptual Guidance (SRPG).} First, we encode target QR code $\mathbf{y}$ and prompt $p$ to embeddings for ControlNet input. In Stage-1, we utilize the pre-trained ControlNet to generate an attractive yet unscannable QR code. In Stage-2, we convert the QR code from Stage-1 into a latent representation $\Tilde{\mathbf{z}}_T$ by adding Gaussian noise and transforming the $\mathbf{y}$ to $\Tilde{\mathbf{y}}$, which has a more similar pattern as $\hat{\mathbf{x}}$, using Qart~\cite{qartcodes}. Finally, we feed the latent and the transformed code into ControlNet, guided by Scanning Robust Perceptual Guidance (SRPG), to create an aesthetic QR code with scannability.}
    \label{fig:two_stage_generation_pipeline}
    \vspace{-15pt}
\end{figure*}

\subsection{Image Diffusion Models}
Recently, Diffusion Models~\cite{sohl2015deep, ho2020denoising} have emerged as powerful generative models, demonstrating superior unconditional image generation capabilities compared to GAN-based models~\cite{goodfellow2020generative, dhariwal2021diffusion}.
Dhariwal et al.~\cite{dhariwal2021diffusion} introduced the concept of Classifier Guidance to control the sampling process via gradients from pre-trained classifiers, which has since been further developed with more generalized conditional probability terms for more freely control~\cite{liu2023more, kim2022diffusionclip, zhao2022egsde, avrahami2022blended, yu2023freedom}.

However, Diffusion Models require substantial computational resources, especially for high-resolution images. To address this issue, Rombach et al.~\cite{rombach2022high} proposed the Latent Diffusion Model (LDM), leveraging a pre-trained VAE to compress high-resolution images into a lower-dimensional latent space. This approach enhances efficiency in the diffusion process while preserving visual quality. For more fine-grained manipulations in downstream tasks, Zhang et al.~\cite{zhang2023adding}, Qin et al.~\cite{qin2023unicontrol}, and Zavadski et al.~\cite{zavadski2023controlnet} proposed adaptation strategies that allow users to fine-tune only the extra output layer instead of the entire model.

These advancements have significantly impacted fields including image editing~\cite{meng2021sdedit, dhariwal2021diffusion, nichol2022glide, couairon2022diffedit, hertz2022prompt, yang2024dynamic, mokady2023null, he2023manifold}, text-to-image synthesis~\cite{rombach2022high, ramesh2022hierarchical, ruiz2023dreambooth}, and commercial product developement, exemplified by DALL-E2~\cite{openai2023dalle2} and Midjourney~\cite{midjourney2023}.

\subsection{Aesthetic QR Codes}
\subsubsection{Non-generative-based Models}
Previous works on aesthetic QR codes have focused on three main techniques: module deformation, module reshuffling, and style transfers. Module-deformation methods, such as Halftone QR codes~\cite{Chu:2013:HQRC}, integrate reference images by deforming and scaling code modules while maintaining scanning robustness. Module-reshuffling, introduced by Qart~\cite{qartcodes}, rearranges code modules using Gaussian-Jordan elimination to align pixel distributions with reference images to ensure decoding accuracy. Image processing techniques have also been developed to enhance visual quality, such as region of interest~\cite{xu2021art}, central saliency~\cite{lin2015efficient}, and global gray values~\cite{xu2019stylized}.  Xu et al.~\cite{xu2019stylized} proposed Stylized aEsthEtic (SEE) QR codes, pioneering style-transfer-based techniques for aesthetic QR codes but encountered visual artifacts from pixel clustering. ArtCoder\cite{su2021artcoder} reduced these artifacts by optimizing style, content, and code losses jointly, although some artifacts remain. Su et al.~\cite{su2021q} further improved aesthetics with the Module-based Deformable Convolutional Mechanism (MDCM). However, these techniques require reference images, which leads to a lack of flexibility and variation.

\subsubsection{Generative-based Models}
With the rise of diffusion-based image manipulation and conditional control techniques, previous works such as QR Diffusion~\cite{qrdiffusion}, QR Code AI Art~\cite{qrcodeai} and QRBTF~\cite{qrbtf2023} have leveraged generative power of diffusion models to create aesthetic QR codes, primarily relying on ControlNet~\cite{zavadski2023controlnet} for guidance. However, more fine-grained guidance that aligns with the inherent mechanisms of QR codes remains unexplored. Another non-open-source method Text2QR~\cite{wu2024text2qr}, introduced a three-stage pipeline that first generates an unscannable QR code using pre-trained ControlNet, followed by their proposed SELR independent of the diffusion sampling process to ensure scannability. However, the diffusion sampling process of Text2QR can not guarantee scannability on its own; our approach aims to fill this gap by designing a training-free, fine-grained guidance integrating scannability criteria into the diffusion sampling process. 

\section{Method}

\begin{figure*}[t]
    \centering
    \includegraphics[width=1\textwidth]{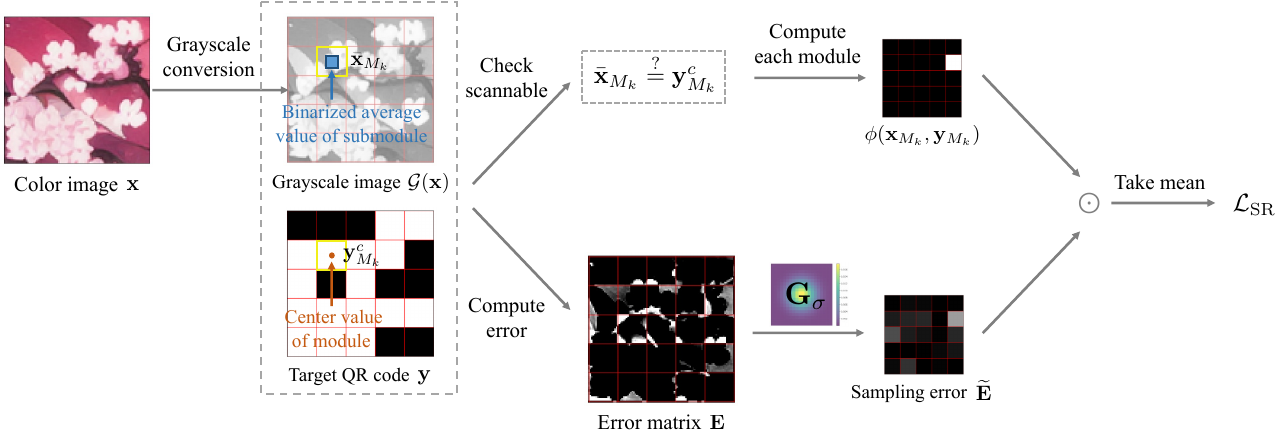}
    \caption{\textbf{An illustration of Scanning Robust Loss (SRL).} SRL is designed at the module level, tailored to the QR code mechanism. The process begins by constructing a pixel-wise error matrix that measures the differences between the pixel values of the target QR code and the grayscale image. Subsequently, the error for each module is re-weighted using a Gaussian kernel, and the central submodule is extracted to implement an early-stopping mechanism. The mechanism stops refining the module and evaluating its error once the average pixel value of the central submodule matches the center pixel value of the target module. Finally, SRL can be calculated as the average error across all modules in the code.}
    \label{fig:srl}
\end{figure*}

DiffQRCoder is designed to generate a scannable and attractive QR code from a given text prompt $p$ and a target QR code $\mathbf{y}$, which consists of $m \times m$ modules, each of size $s \times s$ pixels. 

Fig.~\ref{fig:two_stage_generation_pipeline} illustrates the overall architecture of DiffQRCoder, consisting of two stages. In Stage-1, ControlNet~\cite{zavadski2023controlnet} without our proposed guidance is employed to create a visually pleasing yet unscannable QR code. In Stage-2, we convert the QR code into a latent representation by adding Gaussian noise and transform the $\mathbf{y}$ to $\Tilde{\mathbf{y}}$, which has a more similar pattern as $\hat{\mathbf{x}}$, using Qart~\cite{qartcodes}. We then feed the latent and the transformed code into ControlNet with Scanning Robust Perceptual Guidance (SRPG). The guided loss function includes perceptual loss and our proposed Scanning Robust Loss (SRL), ensuring that the generated QR codes are both scannable and attractive. Besides, we propose a post-processing technique called Scanning Robust Manifold Perceptual Gradient Descent (SR-MPGD) to boost the scanning robustness. Detailed descriptions on SRL, the two-stage pipeline, and SR-MPGD are provided in Sec.~\ref{sec:SRL}, Sec.~\ref{sec:TwoStage}, and Sec.~\ref{sec:SRPGD}, respectively.

\subsection{Scanning Robust Loss}
\label{sec:SRL}
SRL (Fig.~\ref{fig:srl}) is designed to assess the scannability of a beautified QR code with respect to its target code, and aims to provide the guidance signal for image manipulation at the module level.
It begins with an error matrix that evaluates the differences between the pixel values of the target code and the image in grayscale. Next, the matrix is re-weighted to account for varying scanning probabilities across the code. Consequently, we extract the central submodule within each module due to its importance in decoding. Additionally, an early-stopping is implemented to prevent over-optimization.

\paragraph{Pixel-wise Error Matrix.}
Given a normalized image $\mathbf{x}$, a target QR code $\mathbf{y}$, and a grayscale conversion operator $\mathcal{G}$ (cf., Appendix \textcolor{red}{A}). We formulate a pixel-wise error matrix $\mathbf{E}$ that calculates the differences in pixel values between $\mathbf{y}$ and $\mathcal{G}(\mathbf{x})$. $\mathbf{E}$ is calculated as follows:
\begin{equation}
    \begin{aligned}
        \label{eq:sr_error}
        \mathbf{E}=
        & \max(1-2\mathcal{G}(\mathbf{x}), 0)
        \odot
        \mathbf{y} + \\
        & \max(2\mathcal{G}(\mathbf{x})-1, 0)
        \odot
        (1-\mathbf{y}),
    \end{aligned} 
\end{equation}
where $\max(\cdot, \cdot)$ operator is applied pixel-wisely, and $\odot$ denotes the Hadamard product. The first term in the equation addresses the white squares of the QR code, while the second term focuses on the black squares.

\paragraph{Error Re-weighting.}
Not all pixels are equally likely to be scanned. According to ART-UP~\cite{xu2021art}, the scanning probabilities of pixels within each module follow a Gaussian distribution. This implies that pixels closer to the center of each module are more important. Consequently, we re-weight the module error $M_k$ as
\begin{equation}
\widetilde{E}_{M_k} = \sum_{(i, j) \in M_k} \mathbf{G}_\sigma(i, j) \cdot \mathbf{E}(i, j),
\end{equation}
where $(i, j)$ indicates the coordinate of a pixel in $M_k$, and $\boldsymbol{G}_\sigma$ is a Gaussian kernel function with standard deviation $\sigma =\lfloor \frac{s - 1}{5} \rfloor$.

\paragraph{Central Submodule Filter.}
ZXing~\cite{ZXing}, a popular barcode scanning library, notes that only the central pixels of each module are essential for decoding QR codes. According to Chu et al.~\cite{Chu:2013:HQRC}, each module is divided into $3\times3$ submodules. They observed that a QR code remains scannable if the binarized average pixel value in the central submodule matches the center pixel value of the target module.This observation enables the creation of visual variations in the peripheral submodules.

A central submodule filter $\mathbf{F}$ is applied to extract the central submodule:
\begin{equation}
  \label{eq:centralFilter}
    \mathbf{F} = \frac{1}{\lceil \frac{m}{3} \rceil^2}
    \left[
    \begin{array}{ccc}
       \mathbf{O} & \mathbf{O} & \mathbf{O} \\
       \mathbf{O} & \mathbf{I}_{\lceil \frac{m}{3} \rceil \times \lceil \frac{m}{3} \rceil} & \mathbf{O} \\
       \mathbf{O} & \mathbf{O} & \mathbf{O}
    \end{array}
    \right]_{m \times m}.
\end{equation}
The binarized average pixel value of the center submodule $\mathbf{x}_{M_k}$ is calculated as: 
\begin{equation}
    \bar{\mathbf{x}}_{M_k} = \mathbb{I}_{[\frac{1}{2}, 1]}
    \left(
    \sum_{(i, j) \in M_k}
    \mathbf{F}(i, j) \cdot
    \mathcal{G}(\mathbf{x}_{M_k}(i, j))
    \right),
\end{equation}
where $\mathbb{I}_A$ is an indicator function of set $A$.

To determine whether $\mathbf{x}_{M_k}$ is correctly matched, we define the function $\phi$ as:
\begin{equation}
\label{eq:phi}
    \phi(\mathbf{x}_{M_k}, \mathbf{y}_{M_k})=
    \begin{cases}
        0, & \bar{\mathbf{x}}_{M_k} = \mathbf{y}_{M_k}^c, \\
        1, & \bar{\mathbf{x}}_{M_k} \neq \mathbf{y}_{M_k}^c,
    \end{cases},
\end{equation}
where $\mathbf{y}_{M_k}^c$ represents the center pixel value of the target module.

\paragraph{Early-stopping Mechanism.}
We employ an early-stopping mechanism at the module level to prevent over-optimization. This mechanism stops refining a module once it can be correctly decoded, i.e. $\phi(\mathbf{x}_{M_k}, \mathbf{y}_{M_k})=0$.  $\phi$ acts as a switch that determines whether to update $\mathbf{x}_{M_k}$, hence its gradient will not be used to update $\mathbf{x}_{M_k}$. Therefore, we use the stop gradient operator $\operatorname{sg}[\cdot]$ to detach this term from the computation graph. The
 SRL can be expressed as:
\begin{equation}
\label{eq:srl}
\mathcal{L}_\text{SR}(\mathbf{x}, \mathbf{y}) = \frac{1}{N}\sum_{k=1}^{N} \phi (\operatorname{sg}[\mathbf{x}_{M_k}], \mathbf{y}_{M_k}) \cdot \widetilde{E}_{M_k},
\end{equation}
where $N$ is the number of modules.

\subsection{Two-stage Pipeline with Scanning Robust Perceptual Guidance}
\label{sec:TwoStage}
DiffQRCoder utilizes a two-stage pipeline. In Stage-1, we use ControlNet, without our proposed guidance, to create a visually appealing yet unscannable QR code. In Stage-2, we refine the generation process with Scanning Robust Perceptual Guidance (SRPG). This guidance employs a loss function that combines Learned Perceptual Image Path Similarity (LPIPS)~\cite{zhang2018unreasonable}(cf. Appendix \textcolor{red}{B.1}), denote as $\mathcal{L}_\text{LPIPS}$, and $\mathcal{L}_\text{SR}$, ensuring a balance between aesthetics and scanning robustness.

\subsubsection{Stage-1}
In stage-1, we first encode $p$ as the prompt embedding $\mathbf{e}_p$ and $\mathbf{y}$ as the QR code embedding $\mathbf{e}_\text{code}$. And we sample a noise latent $\mathbf{z}_t$ from a standard normal distribution. Then feed into ControlNet to generate an unscannable QR code $\hat{\mathbf{x}}$, which will be used in stage-2 for perceptual regularizing reference.

\subsubsection{Stage-2}
In stage-2 we adopt the unscannable QR code $\hat{\mathbf{x}}$ generated in stage-1 as a starting point for enhancing scannability and a regularizing reference for LPIPS to preserve aesthetics. First, we convert image $\hat{\mathbf{x}}$ into a latent representation $\Tilde{\mathbf{z}}_t$ using the VAE encoder and by adding Gaussian noise. We also transform the target QR code $\mathbf{y}$ to be a more similar pattern as $\hat{\textbf{x}}$ for better conditioning. Both $\Tilde{\mathbf{z}}_t$ and $\Tilde{\mathbf{y}}$ are then fed into ControlNet. The predicted clean latent at each timestep $t$ can be calculated as
\begin{equation}
    \Tilde{\mathbf{z}}_{0|t}=\frac{1}{\sqrt{\bar{\alpha}_t}}\left(\Tilde{\mathbf{z}}_t - \sqrt{1-\bar{\alpha}_t}\epsilon_\theta(\Tilde{\mathbf{z}}_t, t, \mathbf{e}_p, \mathbf{e}_\text{code})\right),
\end{equation}
where $\epsilon_\theta$ denotes the noise predictor of ControlNet.

Since $\mathcal{L}_\text{SR}$ and $\mathcal{L}_\text{LPIPS}$ operate in the pixel space, we use the pre-trained image decoder of ControlNet, $\mathcal{D}_{\theta}(\cdot)$, to map $\Tilde{\mathbf{z}}_{0|t}$ into the pixel space: $\Tilde{\mathbf{x}}_{0|t}=\mathcal{D}_{\theta}
\left(\Tilde{\mathbf{z}}_{0|t}\right)$.
As a result, the guidance function $F_\text{SRP}$ can be formulated as:
\begin{equation}
    \label{eq:F_SRP}
    F_\text{SRP}(\Tilde{\mathbf{z}}_t, \Tilde{\mathbf{y}}, \hat{\mathbf{x}})
    =\lambda_1 \mathcal{L}_\text{SR}(\Tilde{\mathbf{x}}_{0|t}, \Tilde{\mathbf{y}}) + \lambda_2 \mathcal{L}_\text{LPIPS}(\Tilde{\mathbf{x}}_{0|t}, \hat{\mathbf{x}}),
\end{equation}
where $\lambda_1$ and $\lambda_2$ denote the guidance scales.

Song et al.~\cite{song2019generative, song2020score} established a connection between the score function and the estimated noise function, demonstrating that
\begin{equation}
    \epsilon_\theta(\mathbf{z}_t, t, \mathbf{e}_p, \mathbf{e}_\text{code})
    = -{\sqrt{1-\bar{\alpha}_t}}
    \nabla_{\mathbf{z}_t} \log p(\mathbf{z}_{t}).
\end{equation}

Inspired by \cite{dhariwal2021diffusion}, the guided noise prediction becomes:
\begin{equation}
\hat{\epsilon}_t = \epsilon_\theta(\Tilde{\mathbf{z}}_t, t, \mathbf{e}_p, \mathbf{e}_\text{code})+
\sqrt{1-\bar{\alpha}_t}\nabla_{\Tilde{\mathbf{z}}_t}
F_\text{SRP}(\Tilde{\mathbf{z}}_t, \mathbf{y}).
\end{equation}
Finally, we employ the DDIM sampling \cite{song2020denoising}:
\begin{equation}
    \Tilde{\mathbf{z}}_{t-1} =
    \sqrt{\frac{\bar{\alpha}_{t-1}}{\bar{\alpha}_t}} \left(\Tilde{\mathbf{z}}_t-\sqrt{1-\bar{\alpha}_t} \hat{\epsilon}_t \right)
    + \sqrt{1-\bar{\alpha}_{t-1}} \hat{\epsilon}_t.
\end{equation}
After $T$ iterations, we decode $\Tilde{\mathbf{z}}_0$ into the pixel space by $\mathcal{D}_{\theta}(\cdot)$ to obtain the generated aesthetic QR code $\mathbf{x}_0$. The complete algorithm for our two-stage generation pipeline is provided in Appendix \textcolor{red}{C.2}. Detailed derivations of the formulas can be found in Appendix \textcolor{red}{B.2} and \textcolor{red}{B.3}.

\subsection{Post-processing with Scanning-Robust Manifold Projected Gradient Descent (SR-MPGD)}
\label{sec:SRPGD}
SR-MPGD is a post-processing technique proposed to enhance scanning robustness further. Our goal is to minimize $\mathcal{L}_\text{SR}(\mathbf{x}, \mathbf{y})$ while ensuring the refined QR code $\mathbf{x}$ still lies on nature image manifold $\mathcal{M}$. The optimization problem is defined as: $\min_{\mathbf{x} \in \mathcal{M}} \mathcal{L}_\text{SR}(\mathbf{x}, \mathbf{y})$.
This constrained optimization problem can be solved via the Projected Gradient Descent (PGD) algorithm. Inspired by the manifold-preserving nature proposed in \cite{he2023manifold}, we use the pre-trained VAE encoder $\mathcal{E}(\cdot)$ to project the image to its space and then iteratively refine the latent by:
\begin{equation}
\mathbf{z}^{i}_0 = \mathbf{z}^{i-1}_0 - \gamma \nabla_\mathbf{z} \mathcal{L}_\text{SR}(\mathcal{D}(\mathbf{z}^{i-1}_0), \mathbf{y}).
\end{equation}
Note that $\mathbf{z}^{i}_0$ indicates the clean image latent output from Sec. \ref{sec:TwoStage} and in each iterative refinement step, the VAE decoder $\mathcal{D}(\cdot)$ will project the latent back to image manifold.
However, the VAE is imperfect and may introduce reconstruction errors in practice. To mitigate this, we incorporate LPIPS loss as a regularization term with a weight $\lambda > 0$:
\begin{equation}
\label{eq:mpgd_lambda}
\mathcal{L}(\mathbf{x}, \mathbf{y}, \mathbf{x}_0) = \mathcal{L}_\text{SR}(\mathbf{x}, \mathbf{y}) + \lambda \mathcal{L}_\text{LPIPS}(\mathbf{x}, \mathbf{x}_0).
\end{equation}
The rationale for employing LPIPS loss is to facilitate $\mathcal{L}_\text{SR}$ to refine incorrect modules while preserving coarse-grained semantics. Finally, the update rule becomes:
\begin{equation}
\label{eq:mpgd_gamma}
\mathbf{z}^{i}_0 = \mathbf{z}^{i-1}_0 - \gamma \nabla_{\mathbf{z}_{0}} \mathcal{L}(\mathcal{D}(\mathbf{z}^{i-1}_0), \mathbf{y}, \mathbf{x}_0).
\end{equation}
By incorporating this latent optimization, it can converge to local minima against $\mathcal{L}_\text{SR}$ near initial latent $\mathbf{z}_0$.

\section{Experiments}

\subsection{Experimental Settings}

\paragraph{Implementation Details.}
In our experiments, 100 text prompts are generated by GPT-4 as the conditions for Stable Diffusion, consistently using \texttt{easynegative} as the negative prompt. We employ \texttt{Cetus-Mix Whalefall} as the checkpoint for Stable Diffusion~\cite{rombach2022high} and QR code Monster v2~\cite{qrcodemonster2023} as the checkpoint for ControlNet~\cite{zhang2023adding}, with the \texttt{guidance\_scale} set to 1.35 for the latter. We compare with QR code AI Art~\cite{qrcodeai}, QR Diffusion ~\cite{qrdiffusion}, and QRBTF ~\cite{qrbtf2023}. The reason we don't compare with non-generative-based methods is that they rely on reference aesthetic images as input, here we only have prompts describing the aesthetic scenes. And other previous works are unavailable for fair comparison due to non-open-source. Compared models are accessed through their web API using their recommended settings. Detailed parameter settings are provided in Appendix \textcolor{red}{D}.

In our QR code setups, we use Version 3 QR codes configured with a medium (M) error correction level and mask pattern 4. Each code includes an 80-pixel padding, and each module is in the size of $20\times20$ pixels. 
Additionally, the text message \texttt{Thanks reviewer!} is encoded into our QR codes for most experiments in the paper. We also conduct quantitative and qualitative results in different messages for QR code generation. We conduct our experiments using a single NVIDIA RTX 4090 GPU. Generating aesthetic QR codes takes approximately 14 to 18 seconds in our two-stage pipeline, each with 40 inference steps.

\begin{figure*}[t]
    \centering
    \includegraphics[width=0.9\textwidth]{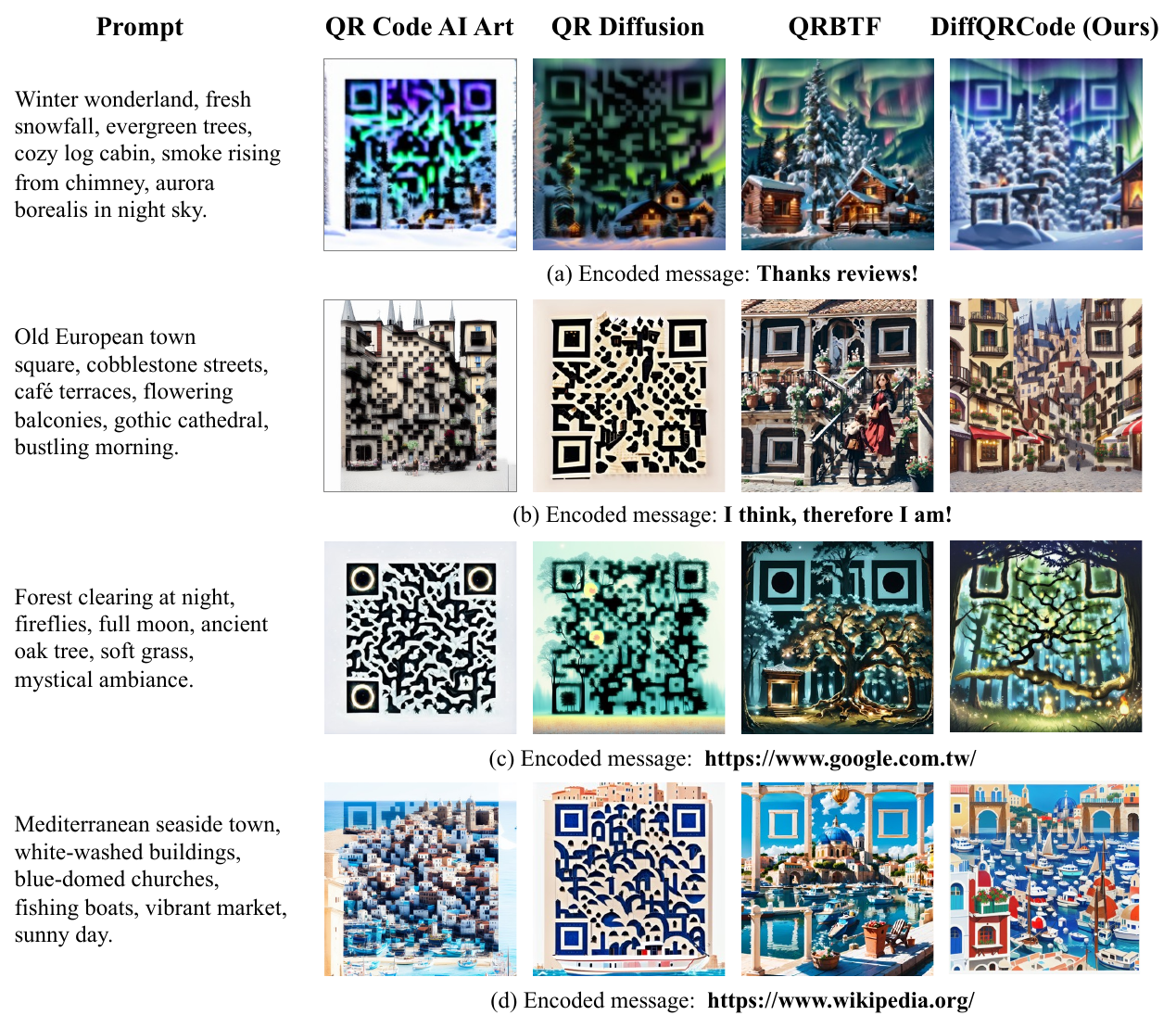}
    \vspace{-10pt}
    \caption{\textbf{Qualitative comparison with other generative-based methods.} DiffQRCoder can generate attractive and scannable QR codes with different encoded messages and prompts.}
    \label{fig:visual_comparisons}
    \vspace{-15pt}
\end{figure*}

\paragraph{Evaluation Metrics.}
We use qr-verify~\cite{fu2023qrverify} to measure Scanning Success Rate (SSR) of aesthetic QR codes. For the quantitative assessment of aesthetics, we use CLIP aesthetic score predictor~\cite{schuhmann2022laion} to reflect image quality and visual appeal.  This score is referred to as the CLIP aesthetic score (CLIP-aes). We also adopt CLIP-score~\cite{radford2021learning} to assess the text-image alignment of generated aesthetic QR codes.

\subsection{Comparison with Other Generative-based Methods}

\vspace{-8pt}
\begin{table}[ht]
    \centering
    \footnotesize{
    \begin{tabular}{lccc}
    \toprule
        \textbf{Method} & \textbf{SSR} $\uparrow$ & \textbf{CLIP-aes.} $\uparrow$ & \textbf{CLIP-score} $\uparrow$ \\
        \midrule
        QR Code AI Art \cite{qrcodeaiart2023} & 90\% & 5.7003 & 0.2341 \\
        QR Diffusion \cite{qrdiffusion} & \second{96\%} & 5.5150 & 0.2780 \\
        QRBTF \cite{qrbtf2023} & 56\% & \first{7.0156} & \first{0.3033} \\
        \midrule
        DiffQRCoder (Ours) & \first{99\%} & \second{6.8233} & \second{0.2992} \\
    \bottomrule
    \end{tabular}
    }
    \vspace{-5pt}
    \caption{Quantitative comparison with other generative-based methods. DiffQRCoder significantly outperforms other methods in SSR with only an insignificant decrease in CLIP-aes.}
    \label{tab:quantitative_comparison}
    \vspace{-25pt}
\end{table}

\paragraph{Quantitative Results.} 
As shown in Tab.~\ref{tab:quantitative_comparison}, we present the quantitative results of our method compared to previous generative-based methods.
Our method outperforms QR Diffusion~\cite{qrdiffusion} and QR Code AI Art~\cite{qrcodeaiart2023} in SSR, CLIP aesthetic score, and CLIP-score. Compared with QRBTF~\cite{qrbtf2023}, our method significantly enhances the SSR, albeit with a little trade-off with CLIP aesthetic score~\cite{schuhmann2022laion}. Notably, the text-image alignment measured by CLIP-score shows that our method is close to QRBTF, indicating our method adheres to prompts without distortion.

Furthermore, we test the robustness of our QR codes under various scenarios, including different simulated scanning angles, error correction level configurations, and scanners from multiple devices and open-source software. As presented in Tab.~\ref{tab:different_angles}, our approach achieves a 97\% SSR even at a $45^{\circ}$ tilt (Implementation details are provided in Appendix \textcolor{red}{D.1}); As presented in Tab.~\ref{tab:different_level} for different error correction levels, our approach still achieves a 96\% SSR under the most rigorous setting (7\% tolerance), we also provide qualitative results in Appendix; As presented in Tab.~\ref{tab:different_device}, for scanning with different scanners, results show that our method can still achieving over 88\% SSR even in the worst case. Additionally, we generate QR codes with various encoded messages and assess their SSR, the results are provided in Tab. \ref{tab:encoded_messages}, and their qualitative results are provided in the Appendix.

\begin{table}[ht]
    \centering
    \small{
    \begin{tabular}{lcccc}
        \toprule
        \textbf{Degree} & $0^\circ$ & $15^\circ$ & $30^\circ$ & $45^\circ$ \\
        \midrule
         \textbf{SSR} $\uparrow$ & 100\% & 100\% & 100\% & 97\% \\
         \bottomrule
    \end{tabular}
    }
    \vspace{-5pt}
    \caption{Scannability of different angles.}
    \label{tab:different_angles}
    \vspace{-15pt}
\end{table}

\begin{table}[ht]
    \centering
    \small{
    \begin{tabular}{lcccc}
        \toprule
        \textbf{Level} & L (7\%) & M (15\%) & Q (25\%) & H (30\%) \\
        \midrule
        \textbf{SSR} $\uparrow$ & 96\% & 100\% & 100\% & 100\% \\
    \bottomrule
    \end{tabular}
    }
    \vspace{-5pt}
    \caption{Scannability of different QR code error correction level.}
    \label{tab:different_level}
    \vspace{-15pt}
\end{table}

\begin{table}[ht]
    \centering
    \small{
    \begin{tabular}{ccccc}
        \toprule
        \textbf{Device} & \textbf{qr-verify} & \textbf{iPhone 13} & \textbf{Pixel 7} \\
        \midrule
        \textbf{SSR $\uparrow$} & 100.00\% & 97\% & 88\% \\
        \bottomrule
    \end{tabular}
    }
    \vspace{-5pt}
    \caption{Scannability results of different devices.}
    \label{tab:different_device}
    \vspace{-10pt}
\end{table}

\vspace{-5pt}
\begin{table}[ht]
    \centering
    \small{
    \begin{tabular}{lc}
        \toprule
        \textbf{Message} & \textbf{SSR} $\uparrow$ \\
        \midrule
        \texttt{I think, therefore I am.} & 97\% \\
        \texttt{You are the apple of my eye.} & 100\% \\
        \texttt{https://www.google.com.tw/} & 100\% \\
        \texttt{https://www.wikipedia.org/} & 97\% \\
    \bottomrule
    \end{tabular}
    }
    \vspace{-5pt}
    \caption{Scannability of different QR code encoded messages.}
    \vspace{-22pt}
    \label{tab:encoded_messages}
\end{table}

\paragraph{Qualitative Results.}
\begin{figure*}[t]
    \centering
    \includegraphics[width=0.90\textwidth]{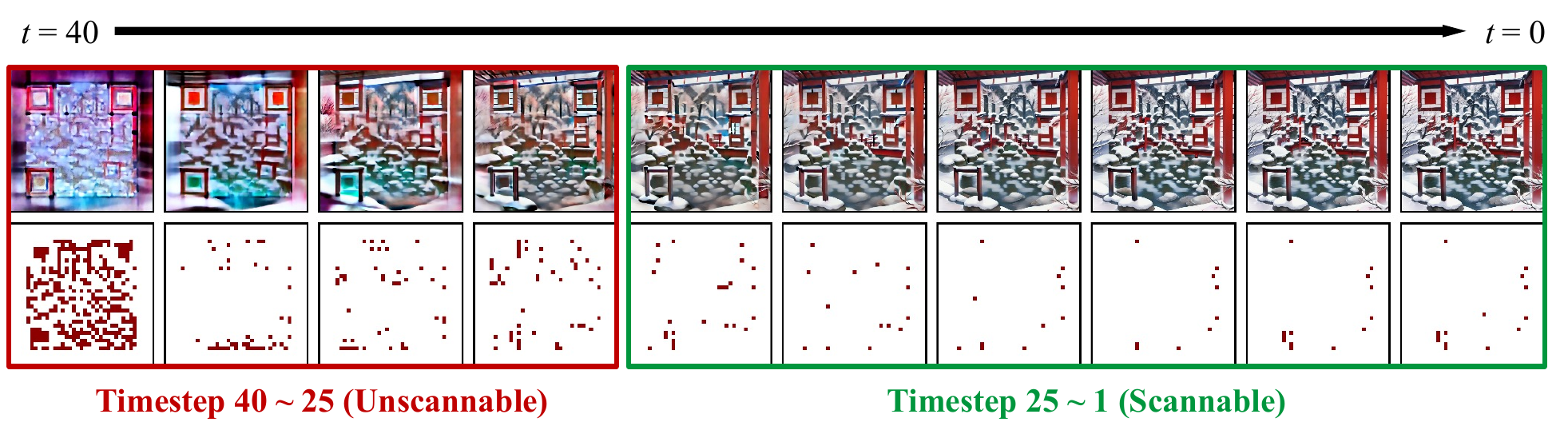}
    \vspace{-10pt}
    \caption{Visualization of $\mathbf{x}_{0 \mid t}$ and its error modules during sampling steps.}
    \label{fig:iterative_refinement_process}
    \vspace{-15pt}
\end{figure*}

We present qualitative comparisons with previous methods in Fig.~\ref{fig:visual_comparisons}. Compared to QR Code AI Art and QR Diffusion, our method exhibits a more harmonized pattern blending with concepts in prompts; Compared to QRBTF, our method trades little aesthetics for scannability, achieving more scanning-robust QR code generation than ControlNet-only methods.
Fig.~\ref{fig:iterative_refinement_process} illustrates the reduction of error modules during the iterative refinement process using our SRPG method. The error modules, highlighted in red, progressively diminish as the denoising step advances. Once the error level falls below a tolerable threshold, the QR code becomes scannable. Moreover, we present qualitative results for a given QR code in different prompts in Fig.~\ref{fig:teaser}. We provide more qualitative results in Appendix.

\paragraph{Subjective Results.}
We conducted a user-subjective aesthetic preference study with 387 participants, the result is reported in Tab.~\ref{tab:user_study}. Although QRBTF~\cite{qrbtf2023} ranked first, our method closely follows with little difference. Considering the limited scannability of QRBTF, which achieved only 56\%, our approach is the leading method for effectively balancing visual attractiveness with scannability. The details of the average rank calculation and questionnaire design are provided in the Appendix \textcolor{red}{E}.

\vspace{-2pt}
\begin{table}[h]
    \centering
    \small{
    \begin{tabular}{lcc}
        \toprule
        \textbf{Methods} & \textbf{Average Rank} $\downarrow$ & \textbf{SSR} $\uparrow$ \\
        \midrule
        QR Code AI Art~\cite{qrcodeaiart2023} & 2.71 & 90\% \\
        QR Diffusion~\cite{qrdiffusion} & 3.18 & \second{96\%} \\
        QRBTF~\cite{qrbtf2023} & \first{1.86} & 56\% \\
        \midrule
        DiffQRCoder (Ours) & \second{2.25} & \first{99\%} \\
        \bottomrule
      \end{tabular}
      }
      \vspace{-5pt}
      \caption{The weighted aesthetic ranks for different methods.}
      \label{tab:user_study}
      \vspace{-13pt}
\end{table}

\begin{table}[t]
    \centering
    \footnotesize{
    \begin{tabular}{lccccc}
    \toprule
        \textbf{Stage} & $\lambda_1$ & $\lambda_2$ & \textbf{SR-MPGD} & \textbf{CLIP-aes.} $\uparrow$ & \textbf{SSR} $\uparrow$ \\
        \midrule
        Stage-1-only & - & -  & & 7.0661 & 60\% \\
        \midrule
        Two-stage & 400 & 0 & & 6.7860 & 86\% \\
        Two-stage & 500 & 0 & & 6.7259 & 88\% \\
        Two-stage & 600 & 0 & & 6.7183 & 94\% \\
        Two-stage & 1000 & 0 & & 6.5667 & 93\% \\
        \hline
        Two-stage & 400 & 0 & \Checkmark & 6.7567 & 98\% \\
        Two-stage & 500 & 0 & \Checkmark & 6.7097 & 100\% \\
        Two-stage & 600 & 0 & \Checkmark & 6.7002 & 99\% \\
        Two-stage & 1000 & 0 & \Checkmark & 6.5629 & 99\% \\
        \hline 
        \hline
        Two-stage & 500 & 2 & & 6.8600 & 90\% \\
        Two-stage & 500 & 3 & & 6.8744 & 89\% \\
        Two-stage & 500 & 5 & & 6.8357 & 89\% \\
        Two-stage & 500 & 10 & & 6.8409 & 88\% \\
        \hline
        Two-stage & 500 & 2 & \Checkmark & 6.8204 & 98\% \\
        \rowcolor{cyan!10}
        Two-stage & 500 & 3 & \Checkmark & 6.8233 & 99\% \\
        Two-stage & 500 & 5 & \Checkmark & 6.7779 & 100\% \\
        Two-stage & 500 & 10 & \Checkmark & 6.8040 & 97\% \\
    \bottomrule
    \end{tabular}
    }
    \vspace{-5pt}
    \caption{Ablations for our proposed pipeline.}
    \label{tab:ablations}
    \vspace{-15pt}
\end{table}

\subsection{Ablation Studies}

\paragraph{Effectiveness of Different SRPG Guidance Scales.}
In this study, we investigate the effectiveness of our proposed $\mathcal{L}_\text{SR}$ and regularizing $\mathcal{L}_\text{LPIPS}$ respectively (cf., Eq.~\ref{eq:F_SRP}). In Tab. \ref{tab:ablations}, Stage-1-only indicates only ControlNet is adopted to generate aesthetic QR codes. First, we fix $\lambda_2 = 0$, and only perform Stage-2 generation with SRPG. In the absence of $\hat{\mathbf{x}}$, we use ControlNet with original $\mathbf{y}$ and text prompts to generate images. As shown in upper half of Tab.~\ref{tab:ablations}, increasing $\lambda_1$ significantly improves SSR while slightly decreasing CLIP aesthetic score. Second, we fix $\lambda_1 = 500$ and perform a full two-stage generation with SRPG. As shown in lower half of Tab.~\ref{tab:ablations}, increasing $\lambda_2$ improves CLIP aesthetic score while preserving SSR.

\paragraph{Effectiveness of SR-MPGD.}
SR-MPGD is a post-processing technique designed to enhance scanning robustness further. In our experiments, we set step size $\gamma = 1000$ (cf., Eq.~\ref{eq:mpgd_gamma}), and LPIPS $\lambda = 0.01$(cf., Eq.~\ref{eq:mpgd_lambda}). As reported in Tab.~\ref{tab:ablations}, it substantially improves SSR with only a negligible decrease in CLIP aesthetic score. By implementing SR-MPGD, we can even achieve 100\% SSR in certain cases.

\section{Conclusion}
In this paper, we introduce a novel training-free Diffusion-based QR Code generator (DiffQRCoder). We propose Scanning-Robust Loss (SRL) to enhance QR code scannability and establish its connection with Scanning-Robust Perceptual Guidance (SRPG). Our two-stage generation pipeline with iterative refinement integrates SRPG to produce aesthetic QR codes. Additionally, we introduce Scanning-Robust Manifold Projected Gradient Descent (SR-MPGD) to further ensure scannability. Compared to existing methods, our approach significantly improves SSR without compromising visual appeal and is competent for real-world applications.

\section*{Acknowledgements}
This research is supported by National Science and Technology Council, Taiwan (R.O.C), under the grant number of NSTC-112-2634-F-002-006, NSTC-112-2222-E-001-001-MY2, NSTC-113-2634-F-001-002-MBK, NSTC-113-2221-E-002-201 and Academia Sinica
under the grant number of AS-CDA-110-M09. We sincerely thank Ernie Chu for his inspiring discussions and valuable feedback.

{
\small
\bibliographystyle{ieee_fullname}
\bibliography{egbib}
}

\clearpage
\renewcommand\thesection{\Alph{section}}
\setcounter{section}{0}
\begin{center}
    \LARGE{\textbf{Appendix}}
\end{center}

\section{Grayscale Conversion}
We denote $\mathcal{G}(\cdot)$ as the grayscale operator, which is defined as:
\begin{equation*}
    \mathcal{G}(\mathbf{x}) = c_r \mathbf{x}^r + c_g \mathbf{x}^g + c_b \mathbf{x}^b,
\end{equation*}
where $\mathbf{x}^r$, $\mathbf{x}^g$ and $\mathbf{x}^b$ are $R$, $G$ and $B$ channels of the image $\mathbf{x}$, respectively. The coefficients $c_r=0.299$, $c_g=0.587$, and $c_b=0.114$ are chosen according to the YCbCr color space standards for grayscale conversion.

\section{Scanning Robust Perceptual Guidance (SRPG)}
\subsection{Learned Perceptual Image Patch Similarity (LPIPS)}
Traditional image-level similarity metrics, which typically compare pixels directly, often fail to align with human perception. To address this issue, Zhang et al.~\cite{zhang2018unreasonable} employed the pre-trained NN-based feature extractors, such as VGG and AlexNet, to transform images into a feature space for comparison. Given our focus on assessing ``aesthetics,'' a high-level and abstract semantic concept, we employ LPIPS for a more appropriate evaluation. LPIPS loss $\mathcal{L}_\text{LPIPS} (\mathbf{x}, \hat{\mathbf{x}}) $ is defined as:
\begin{equation*}
    \mathcal{L}_\text{LPIPS}(\mathbf{x}, \hat{\mathbf{x}}) =
    \sum_{l, i, j} \frac{1}{h_l w_l}
    \| \omega^l \odot (\psi^l(\mathbf{x})_{i, j}-\psi^l(\hat{\mathbf{x}})_{i, j}) \|_2^2,
\end{equation*}
where $\psi^l(\cdot)$ denotes features extracted from the $l$-th layer, $(h_l$, $w_l)$ are the height and width of $\psi^l(\mathbf{x})$, and $\omega^l$ is a channel-wise scaling vector.

\subsection{Derivation of Conditional Probability Term in Generalized Classifier Guidance}
Song et al.~\cite{song2019generative, song2020score} established a connection between the score function $\nabla_{\mathbf{z}_t} \log p(\mathbf{z}_{t})$ and the noise estimation function $\epsilon_\theta(\mathbf{z}_t, t, \mathbf{e}_p, \mathbf{e}_\text{code})$ via Tweedie's Formula \cite{efron2011tweedie}
\begin{equation}
    \epsilon_\theta(\mathbf{z}_t, t, \mathbf{e}_p, \mathbf{e}_\text{code})
    = -{\sqrt{1-\bar{\alpha}_t}}
    \nabla_{\mathbf{z}_t} \log p(\mathbf{z}_{t}).
\end{equation}
Inspired by \cite{dhariwal2021diffusion}, to perform conditional sampling, we substitute the score function with a conditional probability term $p(\mathbf{z}_t | \mathbf{y})$. Then we rewrite the conditional probability term using Bayes' Theorem. Specifically, we define the updated score estimate $\hat{\epsilon}_t$ with condition $\mathbf{y}$ at timestep $t$ as:
\begin{align*}
    \hat{\epsilon}_t
    &:=-{\sqrt{1-\bar{\alpha}_t}}
    \nabla_{\mathbf{z}_t} \log p(\mathbf{z}_t | \mathbf{y}) \\
    &= -{\sqrt{1-\bar{\alpha}_t}}
    \nabla_{\mathbf{z}_t} \log 
    \left( \frac{p(\mathbf{z}_t)p(\mathbf{y}|\mathbf{z}_{t})}{p(\mathbf{y})}\right) \\
    &= -{\sqrt{1-\bar{\alpha}_t}}
    \left(\nabla_{\mathbf{z}_t} \log p(\mathbf{z}_t)
    +\nabla_{\mathbf{z}_t} \log p(\mathbf{y}|\mathbf{z}_{t}) \right) \\
    & = \epsilon_\theta(\mathbf{z}_t, t, \mathbf{e}_p, \mathbf{e}_\text{code})
    -{\sqrt{1-\bar{\alpha}_t}}
    \nabla_{\mathbf{z}_t} \log p(\mathbf{y}|\mathbf{z}_{t}).
\end{align*} 

Following \cite{bansal2024universal}, we define $F$ as the guidance function, thus the final updated estimated score becomes:
\begin{equation}
    \hat{\epsilon}_t = \epsilon_\theta(\mathbf{z}_t, t, \mathbf{e}_p, \mathbf{e}_\text{code})
    + {\sqrt{1-\bar{\alpha}_t}}
    \nabla_{\mathbf{z}_t} F(\mathbf{z}_t, \mathbf{y}).
\end{equation}

\subsection{Derivation of SRPG Gradient}
In this section, we derive the gradient of the guidance function. Given the expression
\begin{equation}
    \Tilde{\mathbf{x}}_{0|t} = \mathcal{D}_{\theta}
    \left(\frac{1}{\sqrt{\bar{\alpha}_t}}\left(\Tilde{\mathbf{z}}_t - \sqrt{1-\bar{\alpha}_t}
    \epsilon_\theta(\Tilde{\mathbf{z}}_t, t, \mathbf{e}_p, \mathbf{e}_\text{code})\right)\right),
\end{equation}
which involves the VAE decoder calculation, we must apply the Chain Rule to derive the gradient. 

Consequently, the gradient of our proposed generalized classifier guidance function $F_\text{SRP}$ can be derived as follows:
\begin{align*}
    &\nabla_{\Tilde{\mathbf{z}}_t} F_\text{SRP}(\Tilde{\mathbf{z}}_t, \Tilde{\mathbf{y}}, \hat{\mathbf{x}})
    =\lambda_1 \nabla_{\Tilde{\mathbf{z}}_t} \mathcal{L}_\text{SR}(\Tilde{\mathbf{x}}_{0|t}, \Tilde{\mathbf{y}}) +
    \lambda_2 \nabla_{\Tilde{\mathbf{z}}_t} \mathcal{L}_\text{LPIPS}(\Tilde{\mathbf{x}}_{0|t}, \hat{\mathbf{x}}) \\
    &= \left(\lambda_1 \frac{\partial \mathcal{L}_\text{SR}(\Tilde{\mathbf{x}}_{0|t}, \Tilde{\mathbf{y}})}{\partial \Tilde{\mathbf{x}}_{0|t}} + \lambda_2 \frac{\partial \mathcal{L}_\text{LPIPS}(\Tilde{\mathbf{x}}_{0|t}, \hat{\mathbf{x}})}{\partial \Tilde{\mathbf{x}}_{0|t}} \right) \cdot
    \frac{\partial \mathcal{D}_{\theta}(\Tilde{\mathbf{z}}_{0|t})}{\partial \Tilde{\mathbf{z}}_{0|t}} \cdot \\
    & \hspace{50pt} \frac{1}{\sqrt{\bar{\alpha}_t}}\left(1 - \sqrt{1-\bar{\alpha}_t} \frac{\partial
    \epsilon_\theta(\Tilde{\mathbf{z}}_t, t, \mathbf{e}_p, \mathbf{e}_\text{code})}{\partial\Tilde{\mathbf{z}}_t}\right).
\end{align*}
where $\hat{\mathbf{x}}$ indicates the reference image generated from Stage-1.

Finally, substitute the conditional score term with $F_\text{SRP}$, the estimated score at timestep $t$ becomes:
\begin{align}
    \hat{\epsilon}_t
    = \epsilon_\theta({\Tilde{\mathbf{z}}_t}, t, \mathbf{e}_p, \mathbf{e}_\text{code})
    +{\sqrt{1-\bar{\alpha}_t}}
    \nabla_{{\Tilde{\mathbf{z}}}_t} F_\text{SRP}({\Tilde{\mathbf{z}}}_t, \Tilde{\mathbf{y}}, \hat{\mathbf{x}}).
\end{align}

\section{Details of Our Proposed Two-stage QR Code Generation Pipeline}

\subsection{Qart}
Qart~\cite{qartcodes} transforms traditional QR codes with user-specified target patterns by exploiting the padding modules. We leverage its capability to create similar patterns of the reference image $\hat{\mathbf{x}}$ from Stage-1 and the target QR code  $\mathbf{y}$, forming a better target QR code $\Tilde{\mathbf{y}}$ for the Stage-2 ControlNet conditioning.

\subsection{Two-stage QR Code Generation Algorithm}
\begin{algorithm}[H]
    \small{
    \caption{Two-stage QR Code Generation Pipeline with Iterative Refinement}
    \label{algo:two_stage_generation_algo}
    \begin{algorithmic}[1] 
        \STATE{\textbf{Input:}
        QR code image $\mathbf{y}$,
        prompt embedding $\mathbf{e}_p$,
        QR code image embedding $\mathbf{e}_\text{code}$,
        UNet $\epsilon_\theta(\cdot, \cdot, \cdot, \cdot)$,
        VAE encoder $\mathcal{E}_{\theta}(\cdot)$
        VAE decoder $\mathcal{D}_{\theta}(\cdot)$,
        sequence $\{\bar{\alpha}_t\}_{t=1}^T$,
        guided weights $\lambda_1, \lambda_2 > 0$,
        error rate $\mathcal{E}(\cdot, \cdot)$,
        and QR code error correction capacity $\tau$.
        }
        \STATE{
        \tikzmk{begin}
        $\mathbf{z}_T \sim \mathcal{N}(\mathbf{0}, \boldsymbol{I})$.
        \hfill $\triangleright$ Stage-1
        }
        \FOR{$t=T$ to $1$}
        \STATE{$\hat{\epsilon} \leftarrow \epsilon_\theta(\mathbf{z}_t, t, \mathbf{e}_p, \mathbf{e}_\text{code})$.}
            \STATE{$\mathbf{z}_{t-1} \leftarrow \sqrt{\frac{\bar{\alpha}_{t-1}}{\bar{\alpha}_t}}\left(\mathbf{z}_t - \sqrt{1-\bar{\alpha}_t} \hat{\epsilon} \right) + \sqrt{1-\bar{\alpha}_{t-1}} \hat{\epsilon}$.}
        \ENDFOR
        \STATE{$\hat{\mathbf{x}} \leftarrow \mathcal{D}_{\theta}(\mathbf{z}_0)$.
        \hfill 
        \tikzmk{end}
        \boxit{algoblue}
        }
        \STATE{$\Tilde{\mathbf{y}} \leftarrow \operatorname{Qart}(\hat{\mathbf{x}}, \mathbf{y})$.
        }
        \STATE{
        \tikzmk{begin}
        $\Tilde{\mathbf{z}}_T \leftarrow \sqrt{\bar{\alpha}_T} \mathcal{E}(\hat{\mathbf{x}}) + \sqrt{1-\bar{\alpha}_T} \epsilon_T, \epsilon_T \sim \mathcal{N}(\mathbf{0}, \boldsymbol{I})$.
        \hfill $\triangleright$ Stage-2
        }
        \FOR{$t=T$ to $1$}     
            \STATE{$\Tilde{\mathbf{z}}_{0 \mid t} \leftarrow \frac{1}{\sqrt{\bar{\alpha}_t}}\left(\Tilde{\mathbf{z}}_t - \sqrt{1-\bar{\alpha}_t} \epsilon_\theta(\Tilde{\mathbf{z}}_t, t, \mathbf{e}_p, \mathbf{e}_\text{code})\right)$.}
            \STATE{$\Tilde{\mathbf{x}}_{0 \mid t} \leftarrow \mathcal{D}_{\theta}(\Tilde{\mathbf{z}}_{0 \mid t})$.}
            \IF{$\mathcal{E}(\Tilde{\mathbf{x}}_{0 \mid t}, \Tilde{\mathbf{y}}) \geq \tau$}
                \STATE{$F_\text{SRP}(\Tilde{\mathbf{z}}_t, \Tilde{\mathbf{y}}, \hat{\mathbf{x}}) \leftarrow \lambda_1 \mathcal{L}_\text{SR}(\Tilde{\mathbf{x}}_{0 \mid t}, \Tilde{\mathbf{y}}) + \lambda_2\mathcal{L}_\text{LPIPS}(\Tilde{\mathbf{x}}_{0 \mid t}, \hat{\mathbf{x}} )$.}
                \STATE{$\hat{\epsilon}_t \leftarrow \epsilon_\theta(\Tilde{\mathbf{z}}_t, t, \mathbf{e}_p, \mathbf{e}_\text{code})+\sqrt{1-\bar{\alpha}_t} \nabla_{\Tilde{\mathbf{z}}_t} F_\text{SRP}(\Tilde{\mathbf{z}}_t, \mathbf{y}, \hat{\mathbf{x}})$.}
            \ELSE
                \STATE{$\hat{\epsilon}_t \leftarrow \epsilon_\theta(\Tilde{\mathbf{z}}_t, t, \mathbf{e}_p, \mathbf{e}_\text{code})$.}
            \ENDIF            \STATE{$\Tilde{\mathbf{z}}_{t-1} \leftarrow \sqrt{\frac{\bar{\alpha}_{t-1}}{\bar{\alpha}_t}}\left(\Tilde{\mathbf{z}}_t - \sqrt{1-\bar{\alpha}_t} \hat{\epsilon} \right) + \sqrt{1-\bar{\alpha}_{t-1}} \hat{\epsilon}$.}
        \ENDFOR
        \STATE{$\mathbf{x}_0 \leftarrow \mathcal{D}_{\theta}(\Tilde{\mathbf{z}}_0)$.
        \hfill
        }
        \tikzmk{end} \boxit{algogreen}
        \RETURN{$\mathbf{x}_0$.}
    \end{algorithmic}
    }
\end{algorithm}

\section{More Details of Experiments}
Our implementation primarily utilizes the \texttt{diffusers} library \cite{von-platen-etal-2022-diffusers} from Hugging Face. Tab.~\ref{tab:different_methods_parameters} outlines the parameters for the various methods used in our experiments; parameters not specified here are set to their default values.

\begin{table}[t]
    \centering
    \begin{tabular}{ll}
        \toprule
        \textbf{Method} & \textbf{Parameters} \\
        \midrule
        QRBTF & Size: 1152px \\
        (Test in June 2024) & Padding ratio: 0.2 \\
        & Anchor style: square \\
        & Correct level: 15\% \\
        \midrule
        QR Code AI Art & ControlNet conditioning scale: 1.1 \\
        & Strength: 0.9 \\
        & Guidance scale: 7.5 \\
        & Sampler: DPM++ Karras SDE \\
        & Seed: 6745177115 \\
        \midrule
        QR Diffusion & QR code weight: 1.65\\
        \midrule
        QR Code Monster & ControlNet conditioning scale: 1.35\\
        \bottomrule
    \end{tabular}
    \caption{Parameter settings in our experiments.}
    \label{tab:different_methods_parameters}
    \vspace{-10pt}
\end{table}

\subsection{Implementation Details of Simulating Scanning Angles}
\begin{figure}[H]
    \centering
    \includegraphics[width=0.8\linewidth]{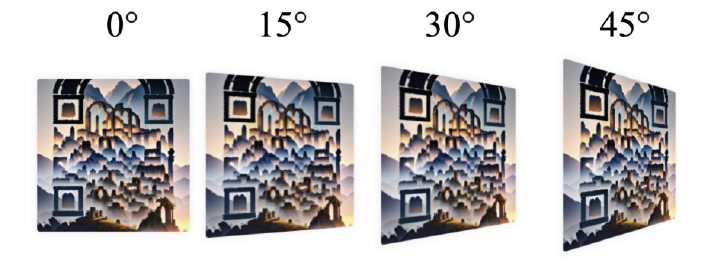}
    \caption{Visualization in different scanning angles of QR code using CSS. Zoom in for better scannability.}
    \label{fig:different_angle}
\end{figure}

The QR codes are randomly chosen from our generated results. These codes are then rotated by 0, 15, 30, and 45 degrees using CSS (Fig.~\ref{fig:different_angle}). A code is considered scannable if it can be scanned within 3 seconds.

\begin{figure*}[t]
    \centering
    \includegraphics[width=0.9\textwidth]{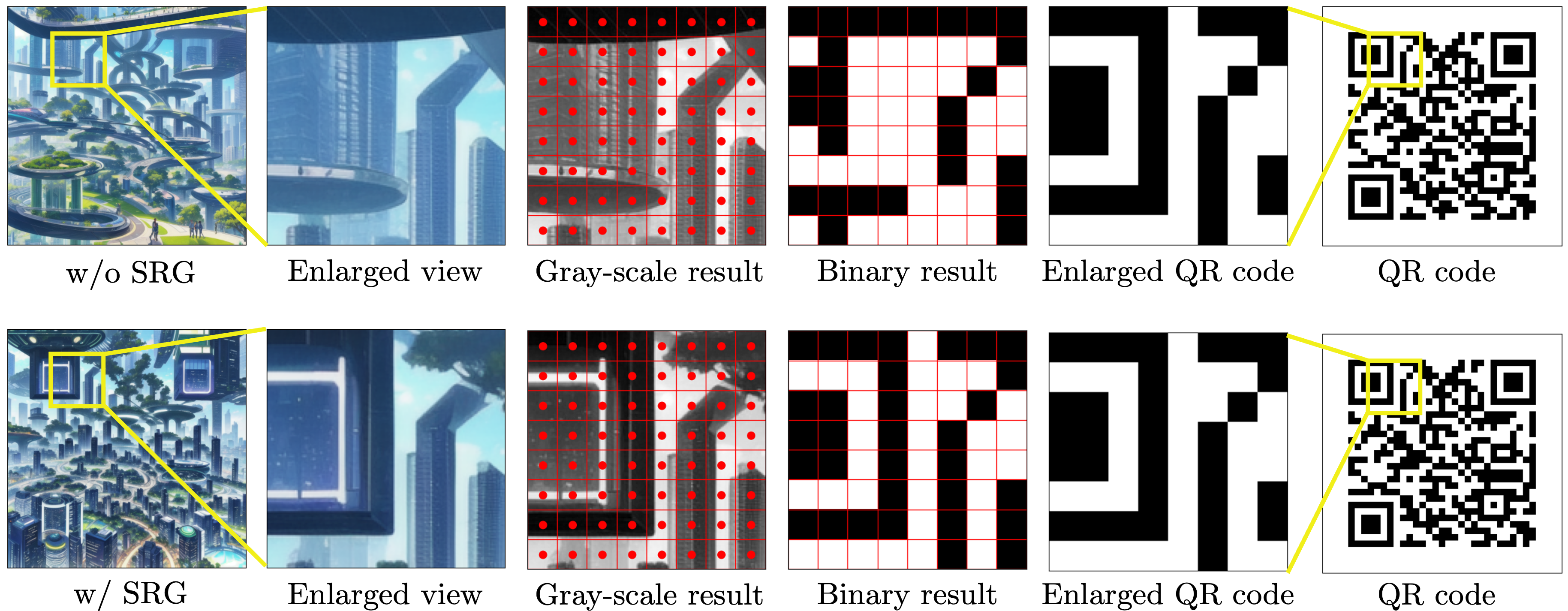}
    \caption{Visual illustration of error analysis.}
    \label{fig:error_analysis}
\end{figure*}

\subsection{Implementation Details of Scanning with Different Scanners}
We chose three widely used QR code scanners for scannability assessment: the built-in scanners on the iPhone 13 and Pixel 7, and the QR Verify software scanner powered by the WeChat decoding algorithm. Our experiment involves scanning 30 aesthetic QR codes ten times for each aesthetic QR code, then calculating the Scanning Success Rate (SSR).

\subsection{Visualization of QR Code Module Error}
We analyze the robustness of the generated results through error analysis. According to SRL, the scanning robustness can be maintained as long as the modules after sampling and binarization yield identical results as the target QR code,  regardless of pixel color changes within these modules.
Fig.~\ref{fig:error_analysis} indicates that our aesthetic QR codes display irregular colors and shapes in their modules. Despite undergoing sampling and binarization, the modules remain consistent with the original QR code. This suggests that our aesthetic QR codes are robust and readable by a standard QR code scanner.

\subsection{Error Analysis}
\subsubsection{Analysis of QR Code Error Rate}
The QR error rate can be computed using the following formula:
\begin{equation}
    \mathcal{E}(\mathbf{x}, \mathbf{y}) = 
    \frac{1}{N}\sum_{k=1}^{N} \phi(\mathbf{x}_{M_k}, \mathbf{y}_{M_k}),
\end{equation} 
where $\mathbf{y}$ is the target QR code, $\mathbf{x}$ is our decoded code, $N$ is the number of modules. The function $\phi(\mathbf{x}_{M_k}, \mathbf{y}_{M_k})$ measures whether the module $M_k$ can be correctly decoded , as defined in Eq. \textcolor{red}{5} in main paper.

As illustrated in Fig.~\ref{fig:qrcode_error}, we set the perceptual guidance scale, $\lambda_2$, to 0 and examine the error rates of a sample across various scanning robust guidance scales, $\lambda_1$, during iterative refinement steps. We observe a marked reduction in error within the first five iterations under our proposed guidance. In contrast, without our guidance, i.e., when $\lambda_1 = 0$, the decrease in error occurs more gradually. Additionally, we visualize QR code errors at different timesteps to better understand the progression of error reduction, the error modules are marked in red, see Fig. \textcolor{red}{6} in main paper. Appendix \textcolor{red}{D.4} further demonstrates how we visualize the error modules.

\subsubsection{Analysis of the Score Magnitude}
\begin{figure}[t]
    \centering
    \begin{subfigure}[b]{0.45\textwidth}
        \includegraphics[width=0.9\textwidth]{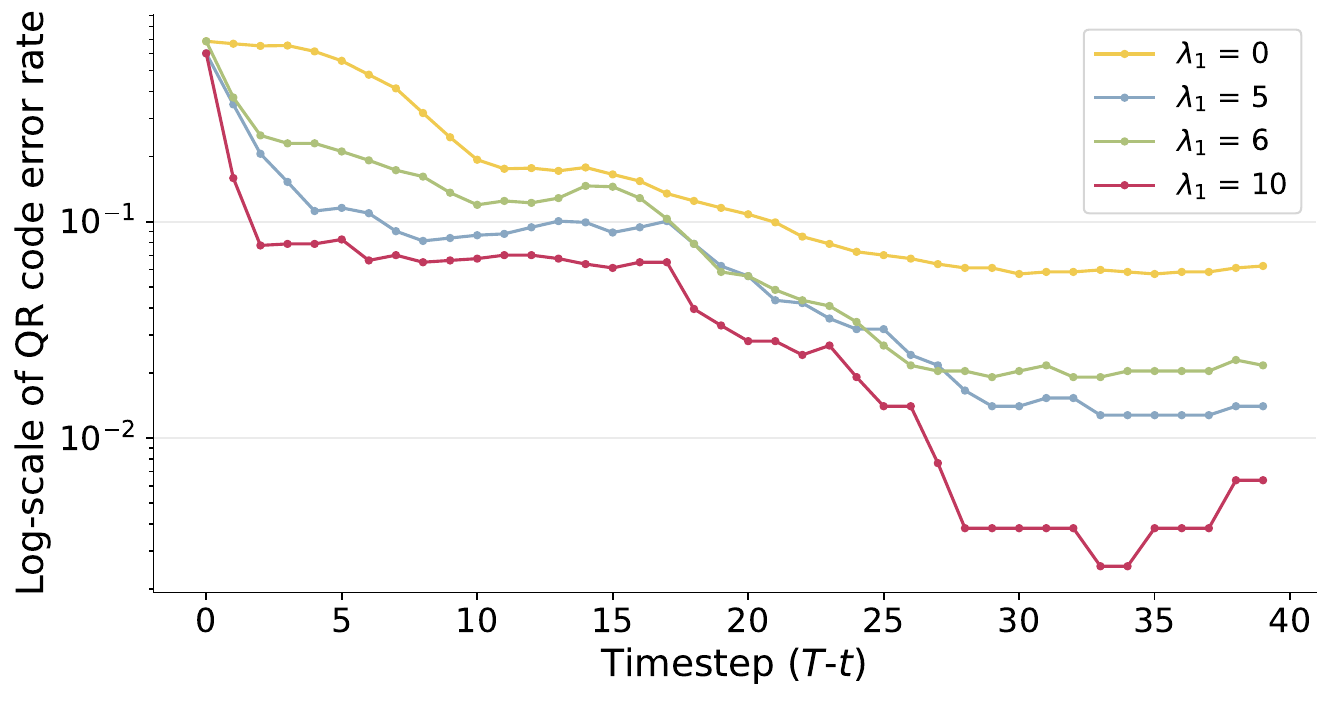}
        \caption{QR code error rate.}
        \label{fig:qrcode_error}
    \end{subfigure}
    \begin{subfigure}[b]{0.45\textwidth}
        \includegraphics[width=0.9\textwidth]{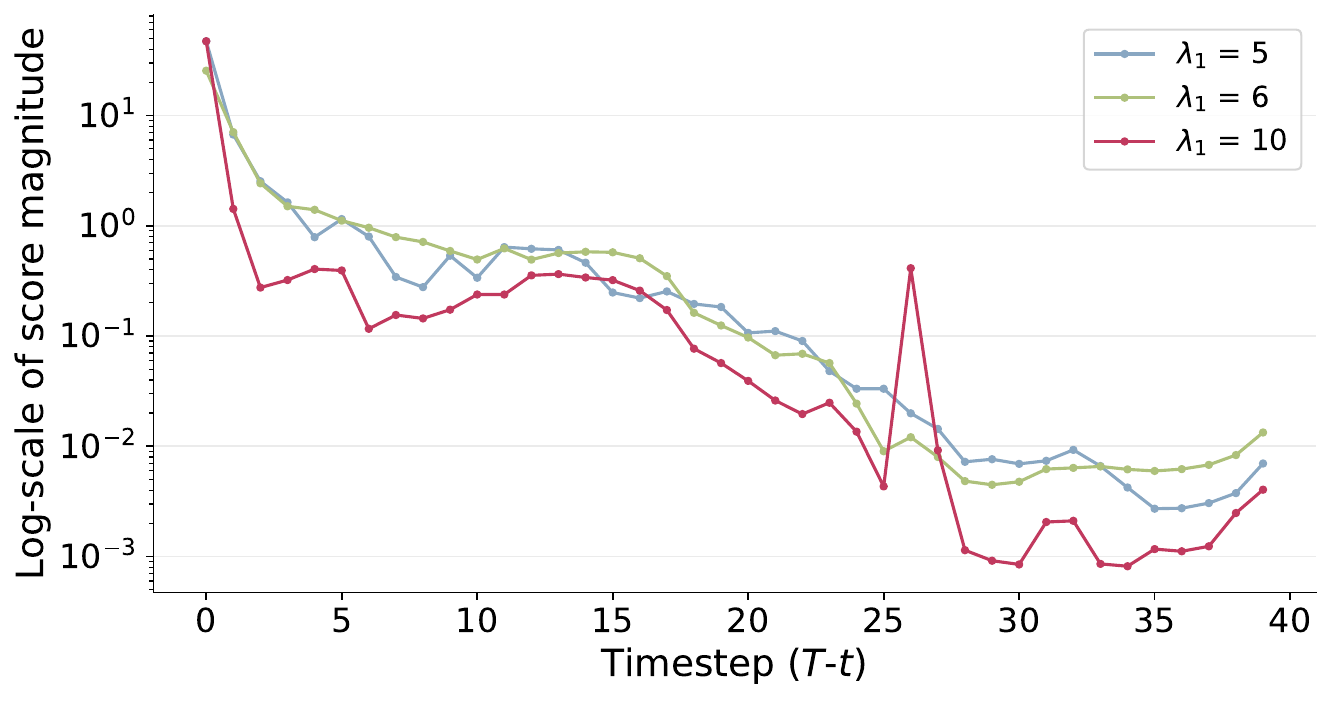}
        \caption{Score magnitude $\|\nabla_{\tilde{\mathbf{z}}_t} F_{\text{SRP}}(\tilde{\mathbf{z}}_t, \mathbf{y})\|_F$.}
        \label{fig:qrcode_score_magnitude}
    \end{subfigure}
    \caption{Error Analysis.}
    \vspace{-10pt}
\end{figure}

Furthermore, we analyze the change in score magnitude $\|\nabla_{\tilde{\mathbf{z}}_t} F_{\text{SRL}}(\tilde{\mathbf{z}}_t, \mathbf{y})\|_F$ across different values of $\lambda_1$. We observe that the score magnitudes decrease over the iterations, suggesting that the effects of guidance diminish over time. This trend is illustrated in Fig.~\ref{fig:qrcode_score_magnitude}. 

\section{User Study}
We conduct a user study with 387 participants. Our subjective test is authorized by the Academia Sinica IRB committee under the approval number AS-IRB-HS 24031.

\subsection{Privacy Issues}
We obtain consent from all participants before they participate in the survey. Additionally, we disclose our data processing policy, which includes the immediate destruction of data after compiling the statistical report, and clarify that no sensitive personal data is collected. Furthermore, participants are informed that they can withdraw from the survey at any time.

\subsection{Question Details}

\begin{figure}
    \centering
    \includegraphics[width=\linewidth]{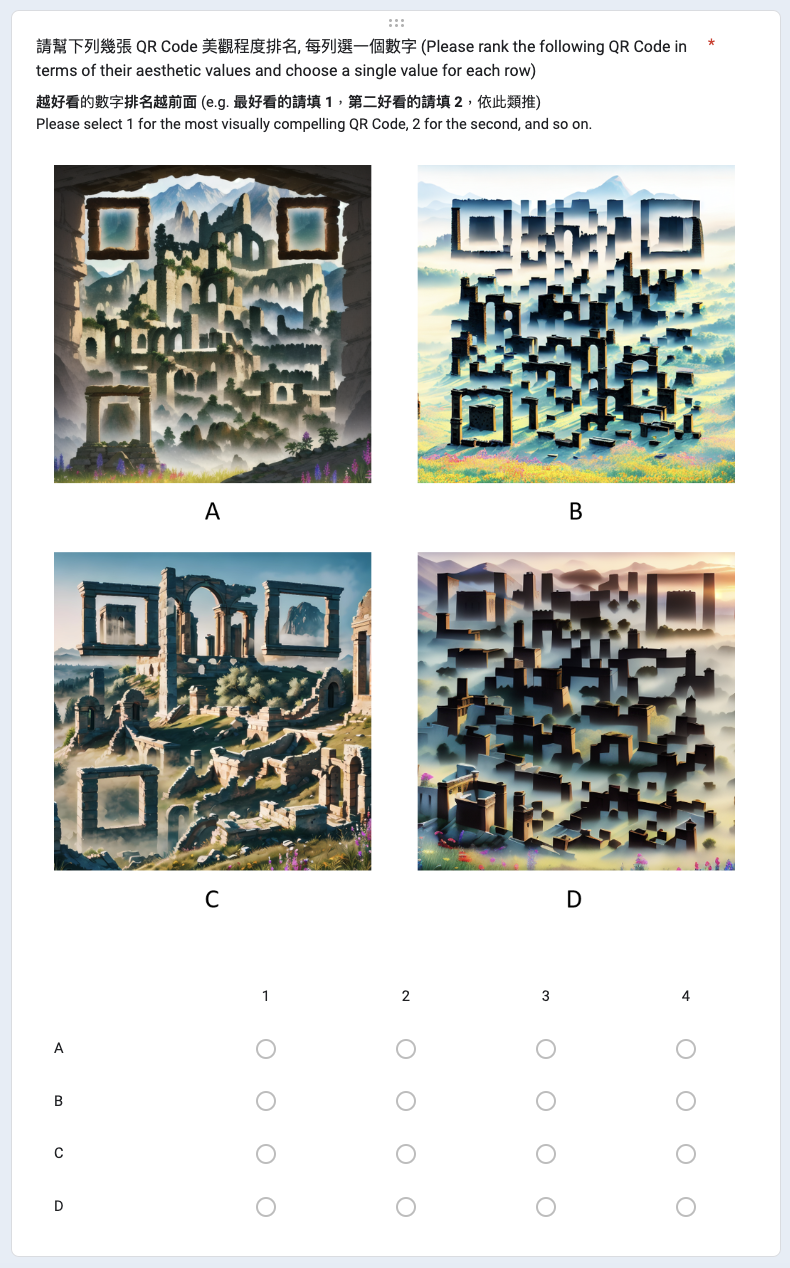}
    \vspace{-10pt}
    \caption{Sample question.}
    \label{fig:user_study}
\end{figure}

Fig.~\ref{fig:user_study} presents a sample of the questions included in our questionnaire, where participants will view four aesthetic QR codes. These codes are generated by QR Diffusion~\cite{qrdiffusion}, QR Code AI Art~\cite{qrcodeaiart2023}, QRBTF~\cite{qrbtf2023}, and our DiffQRCoder. Participants are then asked to rank the options A, B, C, and D based on their perceived aesthetic appeal.

\subsection{Average Ranking Calculation}
To evaluate the results of the user study, we calculate the weighted average rank for each QR code by summing the products of all ranks and their corresponding frequencies, then dividing by the total number of participants. For example, if 20 participants rank a QR code as 1, 10 participants rank it as 2, and 100 participants rank it as 3, the average rank of that QR code can be calculated as follows:
\begin{align*}
\frac{1 \times 20 + 2 \times 10 + 3 \times 100}{130} = 2.615.
\end{align*}

\section{Limitation and Future Work}
Our approach showcases the significant capability of creating aesthetic QR codes, outperforming existing methods. However, it sometimes does not guarantee 100\% scannability and requires hyperparameter adjustments to optimize results. To address this, we apply post-processing to refine our outputs.  Our future work aims at improving the approach into a hyperparameter-insensitive and end-to-end pipeline without post-processing. Additionally, we plan to enhance controllability using image-to-image methodologies to enable more personalized aesthetic QR code generation.

\section{Societal Impacts}
Our proposed approach has potential vulnerabilities, including the risk of being used for phishing, spamming, or disseminating false or inappropriate content. To mitigate these risks, we can implement preventive measures such as URL filtering and prompt blacklisting.

\newpage
\begin{figure*}[t]
    \centering
    \includegraphics[width=0.9\textwidth]{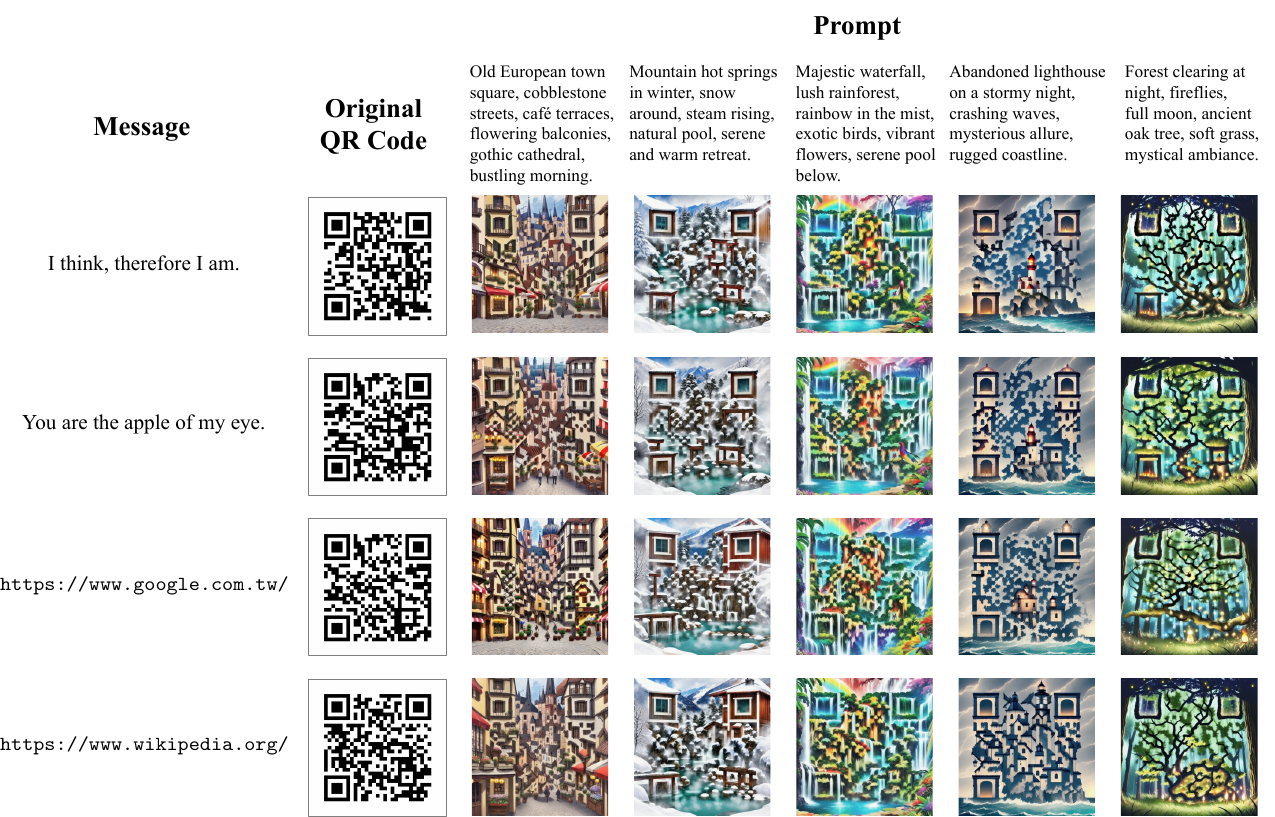}
    \caption{Qualitative results for different QR code messages.}
    \label{fig:different_message}
\end{figure*}

\vspace{-5pt}

\begin{figure*}[t]
    \centering
    \includegraphics[width=0.9\textwidth]{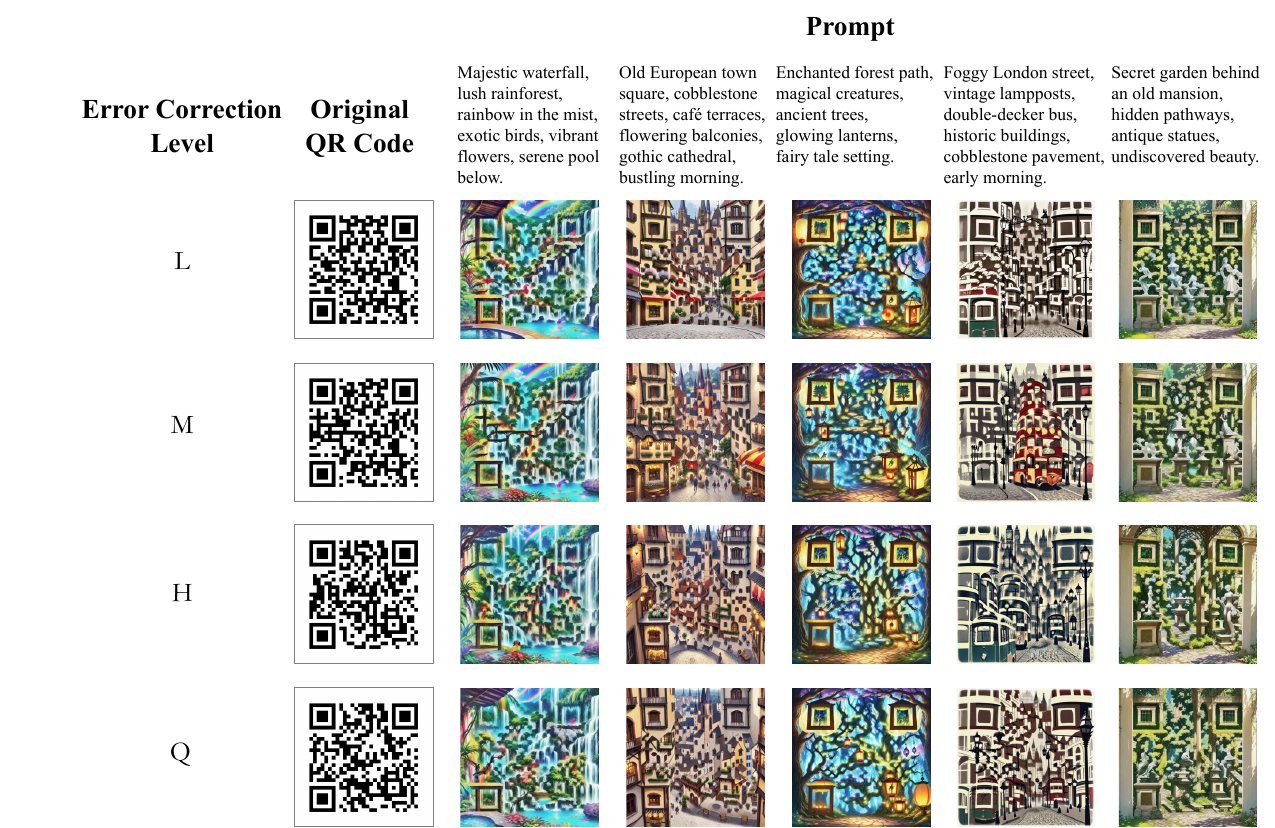}
    \caption{Qualitative results for different QR code error correction levels.}
    \label{fig:different_level}
\end{figure*}

\vspace{-5pt}

\begin{figure*}[t]
    \centering
    \includegraphics[width=0.8\textwidth]{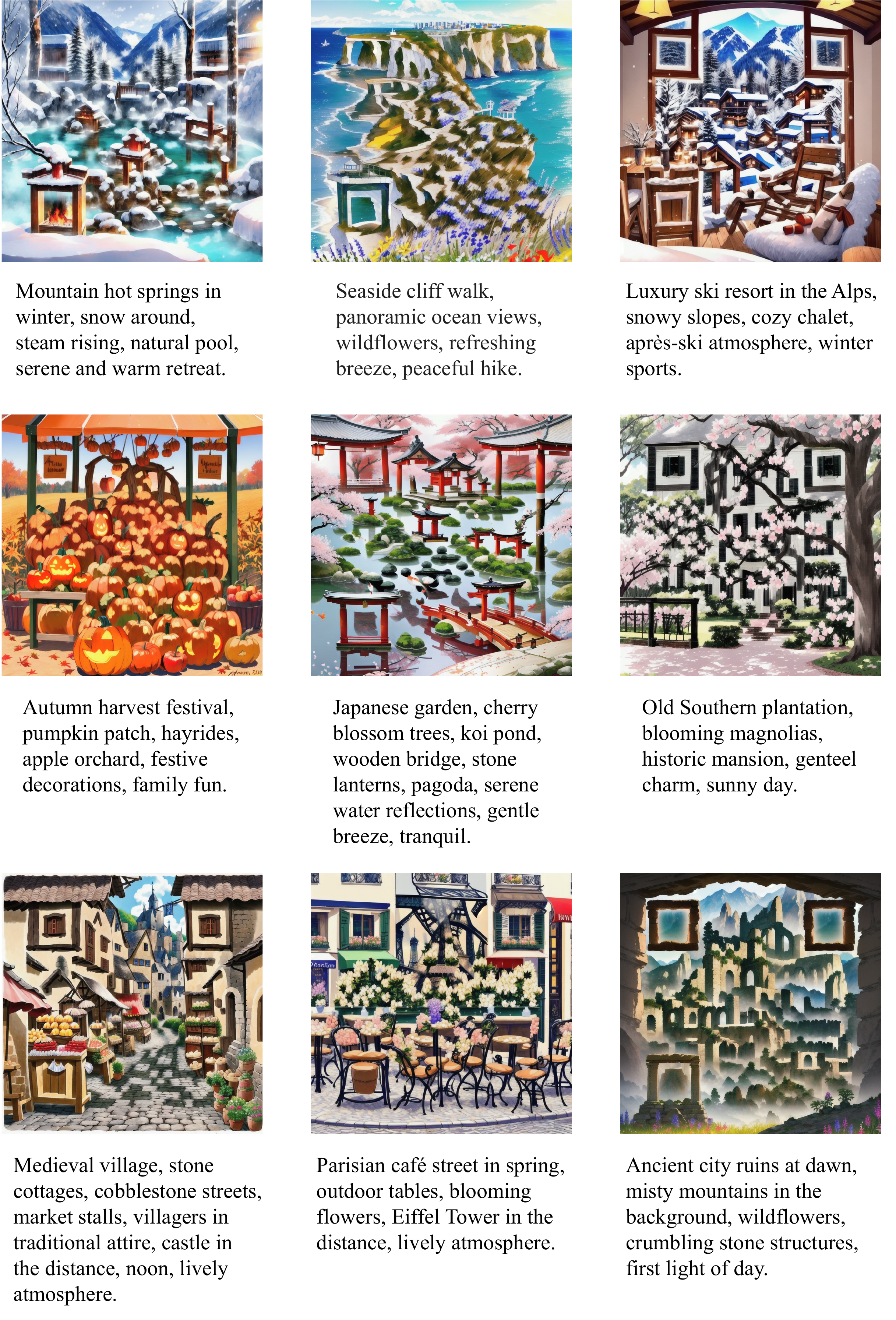}
    \vspace{-10pt}
    \caption{More qualitative results and corresponding prompts.}
    \label{fig:more_result}
\end{figure*}

\end{document}